# Kernel Methods in Remote Sensing: A review

Mahesh Pal
Assistant Professor
Department of Civil Engineering
NIT Kurukshetra, 136119
mpce_pal@yahoo.co.uk

**Abstract:** Kernel-based machine learning algorithms are based on mapping data from the original input feature space to a kernel feature space of higher dimensionality to solve a linear problem in that space. Over the last decade, kernel based classification and regression approaches such as support vector machines have widely been used in remote sensing as well as in various civil engineering applications. In spite of their better performance with different datasets, support vector machines still suffer from shortcomings such as visualization/interpretation of model, choice of kernel and kernel specific parameter as well as the regularization parameter. Relevance vector machines are another kernel based approach being explored for classification and regression with in last few years. The advantages of the relevance vector machines over the support vector machines is the availability of probabilistic predictions, using arbitrary kernel functions and not requiring setting of the regularization parameter. This paper presents a state-of-the-art review of SVM and RVM in remote sensing and provides some details of their use in other civil engineering application also.

## *Introduction*

Support Vector Machines (SVM) is one of the several Kernel-based techniques available in the field of machine learning (Cortes and Vapnik, 1995). It is one of the most sophisticated nonparametric machine learning approach available with various applications and many different configurations (Burges, 1998) depending upon the kernel and optimization method used. SVM provide an innovative means of classification and regression using the principles of structural risk minimization (SRM) which arise from statistical learning theory. Kernel-based methods are based on mapping data from the original input feature space to a kernel feature space of higher dimensionality and solving a linear problem in that space. These methods allow us to interpret and design learning algorithms geometrically in the kernel space (which is nonlinearly related to the input space).

SVM is a good candidate for remote sensing data classification for a number of reasons. Firstly, an SVM can work well with a small training data set as the selection of a sufficient number of pure training pixels has always been a problem with remote sensing data. Secondly, SVM has been found to perform well with high accuracy for problems involving hundreds of features (Gualtieri and Cromp, 1998). The processing of such high-dimensional data to extract quality information has always been a challenge for remote sensing community. Conventional statistical classifiers may fail to process high dimensional data due to the requirement of large training samples. Further, unlike neural networks, SVMs are robust to the overfitting problem as they rely on margin maximization rather than finding a decision boundary directly from the training samples.

Despite the current popularity of SVM, like neural network, this algorithm also require setting of appropriate values for the regularization parameter (C), choice of kernel and kernel specific parameter (such as gamma value with radial basis function kernel). A recent development in the kernel based classification/regression approach is the relevance vector machine (RVM) (Tipping, 2001). The RVM is a Bayesian extension of the SVM with some advantages such as the removal of the need to define the regularization parameter, an ability to use non-Mercer kernels and the providing probabilistic output.

The aim of this article is to discuss the use of SVM/RVM in remote sensing and to discuss about the parameters considered for the design of SVM/RVM. Finally, a brief review of their applications in other civil engineering applications is also provided.

*Support vector machines*

SVM is based on statistical learning theory as proposed by Vapnik and Chervonenkis (1971) and discussed in detail by Vapnik (1995). The SVM can be seen as a new way to train polynomial, radial basis function, or multilayer perceptron classifiers, in which the weights of the network are found by solving a Quadratic Programming (QP) problem with linear inequality and equality constraints using structural risk minimisation rather than by solving a non-convex, unconstrained minimisation problem, as in standard neural network training technique using empirical risk minimisation. Empirical risk minimises the misclassification error on the training set, whereas structural risk minimises the probability of misclassifying a previously unseen data point drawn randomly from a fixed but unknown probability distribution. The name SVM results from the fact that one of the outcomes of the algorithm, in addition to the parameters for the classifiers, is a set of data points (the "support vectors") which contain, in a sense, all the information relevant to the classification problem.

An SVM is basically a linear learning machine based on the principle of optimal separation of classes and was initially designed to solve classification problems. The aim is to find a *linear separating hyperplane* that separates classes of interest. The hyperplane is a plane in a multidimensional space and is also called *a decision surface* or *an optimal separating hyperplane* or *an optimal margin hyperplane*. The linear separating hyperplane is placed between classes in such a way that it satisfies two conditions. First, all data vectors that belong to the same class are placed on the same side of the hyperplane. Second, the distance between the closest data vectors in both the classes is maximized (Vapnik, 1995). In other words, the optimum hyperplane is the one that leaves the maximum margin between the two classes, where margin is defined as the sum of the distances of the hyperplane from the closest data vectors of the two classes. For each class, the data vectors forming the boundary of classes are located on supporting hyperplanes. Thus, these data are called the *Support Vectors* (Schölkopf, 1997) and are the most significant ones for SVM (Schölkopf, 1997).

Many a times, a linear separating hyperplane is not able to classify input data without error. Under such circumstances, the data are transformed to a higher dimensional space using a non-linear transformation function that spreads the data apart such that a linear separating hyperplane may be found in that space. But, due to very large dimensionality of the feature space, it is practically not possible to compute the inner product of two transformed data vectors. This may, however, be achieved by using a *kernel function* in place of the inner product of two transformed data vectors in the feature space. The use of kernel function reduces the computational effort by a significant amount. Other advantages of SVM are their ability to adapt their learning characteristic via a kernel function and an ability to adequately classify data on a high-dimensional feature space with a limited number of training data sets (Vapnik, 1995).

Linearly separable classes are the simplest case on which to train a support vector machine. Let the training data with $k$ number of samples is represented by $\{\mathbf{x_i}, c_i\}$, $i = 1, ..., k$, where $\mathbf{x} \in \mathbf{R}^n$ is an n-dimensional vector and $c \in \{-1, +1\}$ is the class label. These training patterns are said to be linearly separable if a vector $\mathbf{w}$ and a scalar $b$ can be defined so that inequalities (1) and (2) are satisfied.

$$\mathbf{w} \cdot \mathbf{x_i} + b \geq +1 \quad \text{for all } c = +1 \quad (1)$$

$$\mathbf{w} \cdot \mathbf{x_i} + b \leq -1 \quad \text{for all } c = -1 \quad (2)$$

The aim is to find a hyperplane that divides the data so that all the points with the same label are on the same side of the hyperplane. This amount to finding $\mathbf{w}$ and $b$ such that:

$$c_i(\mathbf{w} \cdot \mathbf{x_i} + b) > 0 \quad (3)$$

If a hyperplane exists that satisfies (3), the two classes is said to be linearly separable. In this case, it is always possible to rescale $\mathbf{w}$ and $b$ so that

$$\min_{1 \leq i \leq k} c_i(\mathbf{w} \cdot \mathbf{x_i} + b) \geq 1$$

That is, the distance from the closest point to the hyperplane is $1/\|\mathbf{w}\|$. Then (3) can be written as

$$c_i(\mathbf{w} \cdot \mathbf{x_i} + b) \geq 1 \quad (4)$$

The hyperplane for which the distance to the closest point is maximal is called the *optimal separating hyperplane* (OSH). As the distance to the closest point is $1/\|\mathbf{w}\|$, the OSH can be found by minimising $\|\mathbf{w}\|^2$ under constraint (4). The minimisation procedure uses Lagrange multipliers and Quadratic Programming (QP) optimisation methods. If $\lambda_i$, $i = 1,...,k$ are the non-negative Lagrange multipliers associated with constraint (4), the optimisation problem becomes one of maximising:

$$L(\lambda) = \sum_i \lambda_i - \frac{1}{2} \sum_{i,j} \lambda_i \lambda_j c_i c_j (\mathbf{x_i} \cdot \mathbf{x_j}) \quad (5)$$

under constraints $\lambda_i \geq 0$, $i = 1, .....,k$.

If $\lambda^a = (\lambda_1^a, ......, \lambda_k^a)$ is an optimal solution of the maximisation problem (5) then the optimal separating hyperplane can be expressed as:

$$\mathbf{w}^a = \sum_i c_i \lambda_i^a \mathbf{x}_i \quad (6)$$

The support vectors are the points for which $\lambda_i^a > 0$ when the equality in (4) holds.

If the data are not linearly separable then a slack variable $\xi_i$, $i = 1,.......,k$ can be introduced with $\xi_i \geq 0$ (Cortes and Vapnik, 1995) such that (4) can be written as

$$c_i(\mathbf{w} \cdot \mathbf{x_i} + b) - 1 + \xi_i \geq 0 \quad (7)$$

and the solution to find a generalised OSH, also called a soft margin hyperplane, can be obtained using the conditions

$$\min_{\mathbf{w},b,\xi_1,\ldots,\xi_k} \left[ \frac{1}{2}|\mathbf{w}|^2 + C\sum_{i=1}^{k} \xi_i \right] \quad (8)$$

$$c_i(\mathbf{w}\cdot\mathbf{x}_i + b) - 1 + \xi_i \geq 0 \quad (9)$$

$$\xi_i \geq 0 \quad i = 1, \ldots, k. \quad (10)$$

The first term in (8) is same as in the linearly separable case and control the learning capacity, while the second term controls the number of misclassified points, whereas parameter C is chosen by the user. Larger value of C means assigning a higher penalty to errors.

In situations where it is not possible to have a hyperplane defined by linear equations on the training data, the techniques discussed for linearly separable data can be extended to allow for non-linear decision surfaces. Boser et al., (1992) suggested to map input data into a high dimensional feature space through some nonlinear mapping. The transformation to a higher dimensional space spreads the data out in a way that facilitates the finding of linear hyperplane. After replacing $\mathbf{x}$ by its mapping in the feature space $\Phi(\mathbf{x})$, equation (5) can be written as:

$$L(\lambda) = \sum_i \lambda_i - \frac{1}{2}\sum_{i,j} \lambda_i \lambda_j c_i c_j \left(\Phi(\mathbf{x}_i)\cdot\Phi(\mathbf{x}_j)\right) \quad (11)$$

It is convenient to introduce the concept of *kernel function* K, such that:

$$K(\mathbf{x}_i, \mathbf{x}_j) = \Phi(\mathbf{x}_i)\cdot\Phi(\mathbf{x}_j) \quad (12)$$

A *kernel function* is substituted for the dot product of the transformed vectors. The formulation of the kernel function from the dot product is a special case of *Mercer's theorem* (Vapnik, 1995; Cristianini and Shawe-Taylor, 2000). In this optimisation problem, only the kernel function is computed in place of computing $\Phi(\mathbf{x})$, which could be computationally expensive. By doing this, the training data are moved into a higher-dimensional feature space where the training data may be spread further apart and a larger margin may be found for the optimal hyperplane. For further details of SVM readers are referred to Vapnik (1995) and Cristianini & Shawe-Taylor (2000).

*Relevance Vector Machines*

Relevance vector machine (RVM) is a recent development in kernel based machine learning approaches and can be used as an alternative to SVM for both regression and classification problems. RVM is based on a Bayesian formulation of a linear model with an appropriate prior that results in a sparse representation than that achieved by SVM. RVM is based on a hierarchical prior, where an independent Gaussian prior is defined on the weight parameters in the first level, and an independent Gamma hyper prior is used for the variance parameters in the second level. This results in an overall student-t prior on the weight parameters, which leads to model sparseness (Tipping, 2001). Relevance vector machines (RVM) have recently attracted much interest in various civil engineering applications. RVM can effectively be used for regression and classification problems. Major advantages of RVM over the SVM are:
1. reduced sensitivity to the hyperparameter settings,
2. Ability to use non-Mercer kernels,
3. probabilistic output with fewer relevance vectors for a given dataset and
4. No need to define the parameter C.

Like SVM, the RVM was originally designed for binary classification. In a two class classification by RVM the aim is, essentially, to predict the posterior probability of membership for one of the classes (0 or 1) for a given input x. A case may then be allocated to the class with which it has the greatest likelihood of membership.

For a 2-class problem with training data $\mathbf{X}=(x_1,............,x_n)$ having class labels $C=(c_1,..........,c_n)$ with $c_i \in (-1,1)$. Based on Bernoulli distribution, the likelihood is expected as:

$$p(c/w) = \prod_{i=1}^{n} \sigma\{(y(x_i))\}^{c_i} \left[1 - \sigma\{(y(x_i))\}\right]^{1-c_i} \qquad (13)$$

Where $\sigma(y)$ is the logistic sigmoid function:

$$\sigma(y(\mathbf{x})) = \frac{1}{1+\exp(-y(\mathbf{x}))}$$

To obtain $p(c/w)$, an iterative method has to be used. Let $\alpha_i^*$ denotes the maximum a posteriori estimate of the hyperparameter $\alpha_i$. The maximum a posteriori estimate of the weights ($W_m$) can be obtained by maximizing the following objective function:

$$f(w_1,w_2,..........,w_n) = \sum_{i=1}^{n} \log p(c_i/w_i) + \sum_{i=1}^{n} \log p(w_i/\alpha_i^*) \qquad (14)$$

where the first summation term corresponds to the likelihood of the class labels, and the second term corresponds to the prior on the parameters $w_i$. In the resulting solution, the gradient of $f$ with respect to $w$ is calculated and only those training data having nonzero coefficients $w_i$ (called relevance vectors) will contribute to the decision function. The posterior is approximated around $W_m$ by a Gaussian approximation with

covariance $\qquad \Sigma = -\left(H|_{W_m}\right)^{-1}$

and mean $\qquad \mu = \Sigma \Phi^T B c$

where $H$ is the Hessian of $f$, matrix $\Phi$ has elements $\phi_{ij} = \mathbf{K}(x_i, x_j)$ and B is a diagonal matrix with elements defined by $\sigma(y(x_i))[1-\sigma(y(x_i))]$.

An iterative analysis is followed to find the set of weights that maximizes the function (14) in which the hyperparameters $\alpha_i$, associated with each weight are updated. For further details of RVM, readers are referred to Tipping (2001).

### *SVM/RVM for land cover classification*

The first reported work in the use of SVM for remote sensing data classification was in 1998 (Gualtieri & Cromp, 1998). They used a SVM to classify an Airborne Visible/Infrared Imaging Spectrometer (AVIRIS) image. Zhu and Blumberg (2002) classified an Advanced Spaceborne Thermal Emission and Reflection Radiometer (ASTER) image and reported the results of a classification experiment with an SVM using polynomial and Radial Basis Function (RBF) kernel. Melgani and Bruzzone (2002) used an SVM to classify an AVIRIS image and compared that with K-Nearest

Neighbours (K-NN) and Radial Basis Function neural network (RBF-NN) classifiers. Huang et al. (2002) intensively investigated the accuracy obtained from SVM classifiers on a Landsat TM image. They also reported a better performance by the SVM classifier in comparison with the back propagation neural network, the decision tree and the maximum likelihood classifier. Pal and Mather (2003) reported the use of SVM with multi and hyperspectral data and found that SVM provide an improved performance in comparison to the backpropagation neural network and maximum likelihood classifier with both multi- and hyperspectral data. After 2003, a large number of works reporting the use of SVM for various applications using remote sensing data is reported (Archibald and Fann, 2007; Bazi and Melgani, 2006;Camps-Valls et al., 2004; Camps-Valls and Bruzzone, 2005; Camps-Valls et al., 2006; Camps-Valls et al., 2008; Dixon and Candade, 2008; Dalponte et al., 2008; Foody and Mathur, 2004; 2006, Mathur and Foody, 2008; Melgani and Bruzzone, 2004; Mazzoni et al., 2007; Plaza et al., 2008; Pal and Mather, 2004, 2005, 2006; Pal, 2006; Waske and Benediktsson, 2007).

Like SVM, the RVM may be used for land cover classification problems using remote sensing data. An exhaustive literature survey suggests only two application of RVM for land cover classification. First study by Demir and Erturk (2007) utilized RVM for hyperspectral data classification and compared its performance with SVM. This study suggested that the performance of RVM is slightly inferior to that of SVM with same training dataset but requires a small number of relevance vector in comparison to the support vector used by SVM.

Another study by Foody (2008) used airborne thematic mapper data to classify crop types at an agricultural test site and suggested similar trends in terms of classification accuracy and number of relevance vector as by Demir and Erturk (2007). He also suggested that the probabilistic output information may also be of value to the analyst undertaking the classification especially and may help to identify possible routes to refine the analysis to obtain further increases in classification accuracy.

*Multiclass approaches with SVM and RVM*

Originally, SVM were developed to perform binary classification. However, applications of binary classification are very limited especially in remote sensing where most of the classification problems involve more than two classes. A number of methods to generate multiclass SVM have been proposed in literature and is still a continuing research topic. Most commonly used approaches such as *one vs. one, one vs. rest, Directed Acyclic Graph (DAG), and Error Corrected Output Coding (ECOC)* based multiclass approaches creates many binary classifiers and combines their results to determine the class label of a test pixel. Another category of multi class approach is to modify the binary class objective function and allows simultaneous computation of multiclass classification by solving a single optimisation problem. A survey of remote sensing literature suggest that majority of published work used *one vs. one* or *one vs. rest* multiclass approaches. Study by Pal (2005) examined six approaches for the solution of multiclass classification problem using remote sensing data with SVM. Results from this study suggest the suitability of *one against one approach* for the dataset used in term of classification accuracy and the computational cost. A recent study by Mathur and Foody (2008) suggested that multiclass approach as proposed by Westin and Watkins (1998) works well in comparison to *one vs. one* or *one vs. rest* approaches for their dataset. For RVM,

Demir and Ertürk (2007) used *one vs. one* approach whereas Foody (2008) employed the approach based on the principles of multinomial logistic regression as proposed by Zhang and Malik (2005).

Within last few years a number of work suggesting the use of different multiclass approaches are reported in literature. Some of the multiclass approaches such as dendrogram based approach (Benabdeslem and Bennani, 2006), Fuzzy support vector machines (Abe and Inoue, 2002) was found performing well in comparison to *one vs. one* approach. Some other approaches such as LaRank (Bordes et al., 2007), Half-agaianst-Half (Lei and Govindaraju, 2005) and the use of inter-class confusion (Godbole et al., 2002) have also been proposed for multiclass SVM.

*Choice of Kernel*

The choice of a proper kernel function plays an important role in SVM based classification/regression. A number of kernels are discussed in the literature (Vapnik, 1995), but it is difficult to choose one which gives the best generalization for a given dataset. Pal (2002) used five different kernels (the linear kernel, the polynomial kernel, the Radial Basis Function (RBF) kernel, linear spline and the sigmoid kernel) in order to investigate the effect of kernel choice on classification accuracy using multispectral data and suggested that the radial basis and the linear splines perform equally well and achieve the highest accuracy for this data set. Recently, Camps-Valls et al., (2006) and Camps-Valls et al., (2008) proposed to use new kernels that takes care of spatial and spectral characteristics of remote sensing data for land cover classification and found them to be working well in comparison to RBF kernel. Mercier and Lennon (2003) and Sap and Kohram (2008) used a modified kernel that take into consideration the spectral similarity between support vectors for classification of hyperspectral data. This kernel is based on the use of spectral angle to evaluate the distance between support vectors. Results from these studies suggest that this approach can compete with existing classification methods. Recently, Üstün et al. (2006) proposed a universal Pearson VII function based kernel (PUK) to solve SVM based regression problems. They suggested that this kernel can be an alternative to the linear, polynomial and radial basis function kernels. Studies using remote sensing as well as some other datasets suggests an improved performance by PUK kernel based SVM in comparison with RBF kernel based SVM (Table 1).

Table 1

| Data set | RBF based support vector machine | PUK based support vector machine |
|---|---|---|
| ETM+ data as reported in Pal (2008) | Accuracy =88.66% | Accuracy = 89.59% |
| Data as used in Pal and Deswal (2008) | CC = 0.969<br>RMSE = 0.328 | CC = 0.969<br>RMSE = 0.327 |
| Dataset used in Pal and Goel (2007) | CC = 0.996<br>RMSE = 0.0014 | CC = 0.997<br>RMSE = 0.0013 |

*Parameter selection with SVM*

SVM is a powerful machine learning method for both classification and regression. However, SVM require the user to set two or more parameters which affect the training process as well as their setting can have a profound affect on the final performance and finding good parameters can become a computational burden as the size of the dataset increases. Most of the studies in remote sensing used grid search method together with k-fold cross-validation error for parameter selection with SVM. A grid search tries values of each parameter across the specified search range using geometric steps. Grid searches are computationally expensive because the model must be evaluated at many points within the grid for each parameter. Recently, Yan et al., (2007) proposes an evolutionary algorithm to automatically determine the optimal parameters of SVM with the better classification accuracy and generalization ability simultaneously. They used evolutionary SVM for Land-cover classification using remote sensing data. Their study suggests that the use of evolutionary algorithm for finding the optimal parameters results in improvement in overall accuracy and generalization ability in comparison to a conventional SVM.

Several other approaches using evolution strategies, genetic algorithm, particle swarm optimization and their combination with grid search are proposed in literature (Keerthi, 2002; Frohlich and Zell, 2005; Friedrichs and Igel, 2005; Lorena and de Carvalho, 2006; Liu et al., 2006; Chunhong and Licheng, 2004; Guo et al., 2006) and found performing well in comparison to grid search method for different datasets.

*Visualization of SVM model*

SVM is found to build accurate models in comparison to other popular machine learning approach but these algorithms are at a disadvantage in terms of intuitive presentation of the classifier, particularly when compared to some other supervised learning techniques like classification trees. Achieving high prediction accuracy and the interpretability of models are two important criteria for machine learning algorithms. While high accuracy classifier such as SVM has intensively been explored, its interpretability still poses a difficult problem. A SVM provide only the support vectors, as a subset of training data that defines the decision boundary. This only reduces the number of data to consider in the interpretation but does not give any idea about "which are the most important factors that determine the class of the instance?", and "What is the magnitude of the effect of these?", and "How do various factors interact?" (Jakulin et al., 2005). With in last few years, several work reported various approaches to visualize SVM models. Poulet (2004) proposed a cooperative approach using SVM and visualization methods, whereas Jakulin et al. (2005) used nomograms to visualize SVM models. Nomograms are an established model visualization technique that can graphically encode the complete model on a single page. In another study, Caragea et al. (2004) used tour-based methods to plot aspects of the SVM classifier. Their approach helps in providing insights about the cluster structure in the data, the nature of boundaries between clusters, problematic outliers and suggested that tours can be used to assess the variable importance. Recently, Ustün et al. (2007) proposed a technique to visualize the information content of the kernel matrix and a way to interpret the ingredients of the Support Vector Regression (SVR) model. Their approach gives an idea about what is happening in the complex SVM and which variables of the input space are the most important ones.

*SVM and RVM based regression for Remote sensing Data*

In the field of regression, only few applications of support vector regression (SVR) and RVM based methods are published using remote sensing data. Zhan et al., (2003) used SVR as the nonlinear transfer function between oceanic chlorophyll concentration and marine reflectance and found it working well in comparison to a neural network. Camps-Valls et al., (2006) used the $\varepsilon$-Huber loss function in the SVR formulation for the estimation of biophysical parameters extracted from remotely sensed data and compared it to other cost functions in the SVR, neural networks and classical bio-optical models for the particular case of the estimation of ocean chlorophyll concentration from satellite remote sensing data. They suggested that the proposed model provides more accurate, less biased, and improved results. Wang et al. (2007) used a kernel-based bidirectional reflection distribution function model inversion method for land surface parameter retrieval. Durbha et al., (2007) used SVR for the retrieval of leaf area index (LAI) from multiangle imaging spectroradiometer data.

Camps-Valls et al., (2006) propose a novel formulation of the RVM that incorporates *prior* knowledge of the problem and successfully tested for quantifying chlorophyll-a concentrations accurately from multispectral satellite ocean colour data. They suggested that their approach provides better results than standard RVM formulations, SVR and neural networks.

*SVM and RVM for civil engineering applications*

Soft computing technique, like artificial neural network is being used to solve various problems related to the geotechnical, water resource, transportation and structural engineering in civil engineering. Results obtained by neural network were compared with different equations proposed in literature and found to be working well in comparison to the empirical relations. A neural network based modelling algorithm requires setting up of different learning parameters (like learning rate, momentum), the optimal number of nodes in the hidden layer and the number of hidden layers. A large number of training iterations may force a neural network to overtrain, which may affect the predicting capabilities of the model. Further, presence of local minima is another problem with the use of a back propagation neural network.

Within last few years, a number of research papers have reported the use of support vector machines in civil engineering and found it to be working well in comparison to neural network approach. Works reporting the use of SVM in geotechnical engineering and found it quite effective in comparison to neural network and empirical relations (Pal, 2006; Gill et al., 2006; Gill et al., 2007; Goh and Goh, 2007; Samui, 2008(a), 2008(b), 2008(c); Samui et al., 2008; Zho, 2008). A recent study by Pal and Deswal (2008) suggest an improved performance by Generalized Regression Neural Network (GRNN) in comparison to SVM, thus suggesting the need to further explore the potential GRNN for geotechnical engineering problems. Samui (2007; 2008d) explored the potential of RVM to assess the Seismic liquefaction potential and settlement of shallow foundation on cohesionless soils respectively and found it to be working well in comparison to neural network approach.

Several studies examined the potential of support vector machines in different water resource engineering applications also (Dibike et al., 2001; Liong and Sivapragasam, 2002; Cannas et al., 2004; Sivapragasam and Muttil, 2005; Pal and Goel, 2006; Khan and

Coulibaly, 2006; Pao-shan et al., 2006; Pal and Goel, 2007; Msiza et al., 2007; Hong and Pai, 2007; Singh et al., 2008; Deswal and Pal, 2008; Hanbay et al., and Goel and Pal (in press)). The SVM showed good performance and is proved to be competitive with neural network with different problems related to water resource engineering whereas the study by Msiza et al., (2007) suggested a better performance by ANN approach in water demand forecasting. An exhaustive literature review suggests few application of RVM in water resources engineering (Tripathi and Govindaraju, 2007; Ghosh and Mujumdar, 2008). Ghosh and Mujumdar (2008) reported a better performance by RVM for their dataset in comparison to SVM based regression approach.

As compared to geotechnical and water resource engineering, few works reporting the use of SVM/RVM in other areas of civil engineering are found in literature. An et al. (2007) compared the performance of SVM in conceptual cost estimation in comparison to discriminant analysis model. They suggested a better performance by SVM based approach. Park et al., (2008) used a two-step SVM classifier for railroad track damage identification and suggested that damage estimation rate of 96.67% was achieved by SVM based classifier. Bulut et al., (2004) used SVM for real-time nondestructive structural health monitoring whereas Oh (2007) reported the use of RVM in earthquake engineering and structural health monitoring

*Future Perspectives*

In this article, we reviewed two kernel based classification and regression approach as well as the factor affecting them for remote sensing and civil engineering applications. Except SVM/RVM, nonparametric models such as Bayesian framework for Gaussian processes (GP) models has recently received significant attention in the filed of machine learning. In comparison to backpropagation neural networks, GP are conceptually simpler to understand and implement in practice (Seeger et al., 2003). GP models are closely related to approaches such as support vector machines (Vapnik, 1995) and specially relevance vector machines (Qüinonero-Candela and Rasmussen, 2005) due to the use of kernel functions (Bishop and Tipping, 2003). Compared with SVM based regression model, GP models are probabilistic in nature (Rasmussen and Williams, 2006). The probabilistic models means that it provides reliability of response of GP models on the given input data (Yuan et al. 2008). GP models were initially designed to solve regression problems only but several approaches are proposed to use it for classification (Williams and Barber, 1998; Seeger and Jordan, 2004; Girolami and Rogers, 2006). Due to its better performance in practice, GP models are being applied to solve various problems effectively (Chen et al. 2007; Likar and Kocijan, 2007; Yuan et al. 2008) but so far no study reported its use in civil engineering. Recently, Pal and Deswal (under communication) reported the use of GP regression approach to predict the load capacity of driven piles and found it performing well in comparison to SVM and GRNN for the dataset used in this study (Table 2). Thus, suggesting the data dependent nature of different machine learning algorithms.

Table 2

| Approach | Root mean square error | $R^2$ value (Coefficient of determination) |
|---|---|---|
| Gaussian Process Regression with PUK kernel | 312.18 | 0.898 |
| Gaussian Process Regression with RBF kernel | 335.85 | 0.881 |
| SVMs with PUK kernel | 334.05 | 0.887 |
| SVMs with RBF kernel | 371.89 | 0.855 |
| GRNN | 436.42 | 0.835 |